\documentclass{article} 
\usepackage{aloe2022,times}

\usepackage{amsmath,amsfonts,bm}

\def\eqref#1{equation~\ref{#1}}

\def\1{\bm{1}}

\DeclareMathAlphabet{\mathsfit}{\encodingdefault}{\sfdefault}{m}{sl}
\SetMathAlphabet{\mathsfit}{bold}{\encodingdefault}{\sfdefault}{bx}{n}

\DeclareMathOperator*{\argmax}{arg\,max}
\DeclareMathOperator*{\argmin}{arg\,min}

\usepackage{graphicx}
\usepackage{hyperref}
\usepackage{url}
\usepackage{wrapfig}
\usepackage{amsmath}
\usepackage{subfig}
\usepackage{hyperref}

\usepackage{dsfont}

\title{Watts: Infrastructure for Open-Ended \\ Learning}
\iclrfinalcopy
\author{Aaron Dharna \And Charlie Summers \And Rohin Dasari \And Julian Togelius \And Amy K. Hoover}

\author{Aaron Dharna \\
New Jersey Institute of Technology \\
\texttt{aadharna@gmail.com}\\
\And
Charlie Summers \\
Columbia University \\
\texttt{cgs2161@columbia.edu}\\
\And
Rohin  Dasari\\
New York University \\
\texttt{rd2893@nyu.edu}\\
\And
Julian Togelius \\
New York University \\
\texttt{julian@togelius.com}\\
\And
Amy K. Hoover \\
New Jersey Institute of Technology \\
\texttt{ahoover@njit.edu}\\
}

\begin{document}

\maketitle

\begin{abstract}

This paper proposes a framework called Watts for implementing, comparing, and recombining open-ended learning (OEL) algorithms. Motivated by modularity and algorithmic flexibility, Watts atomizes the components of OEL systems to promote the study of and direct comparisons between approaches. Examining implementations of three OEL algorithms, the paper introduces the modules of the framework. The hope is for Watts to enable benchmarking and to explore new types of OEL algorithms. The repo is available at \url{https://github.com/aadharna/watts} 

\end{abstract}

\section{Introduction}

One approach to designing systems that continue to produce interesting or novel forms (i.e., open-ended evolution or learning; (OEL)) \citep{STANDISH2003, openendedlearningteam2021openended, lehman2011abandoning,  schmidhuberpowerplay2012} is co-evolving agents and the problems that they are trying to solve. The paired open-ended trailblazer algorithm (POET) \citet{wang:gecco19,wang2020enhanced,dharna2020cogeneration} co-evolves bipedal agents and their environments, represented as terrains that the agents must learn to navigate. Through open-ended evolution, environments are proposed that POET-agents can solve more efficiently than agents trained through standard optimization techniques like Proximal Policy Optimization (PPO) \citep{schulman2017proximal}. In PAIRED \citep{dennis2020emergent} game-playing and game-building agents co-optimize together creating a curriculum of 2D mazes resulting in agents that generalize better than agents trained with Minimax or Domain Randomization techniques for curriculum generation. \citet{openendedlearningteam2021openended} recently proposed some methods where a single artificial agent can solve 37 games it has never seen before with behavior that generalizes even to held-out tasks.

While measures aim to quantify the open-endedness of such systems \citep{wang2020enhanced}, at a minimum directly comparing the algorithms first requires testing them in the same domain. Bipedal agents in POET navigate 2-D terrains in the Hardcore BipedalWalker-v3 environment, where stumps, gaps, and stairs create obstacles for the agent to maneuver around \citep{brockman2016openai}. PAIRED \citep{dennis2020emergent} trains agents to solve 2-D mazes from the MiniGrid suite \citep{gym_minigrid} and DeepMind trains agents to solve game rules and environment topologies in their proprietary XLand Universe \citet{openendedlearningteam2021openended}. Decoupling these algorithms from their environments is often challenging. Beyond that, research code is hard to read which can lead to algorithmic misunderstandings \citep{dharna2020cogeneration, dennis2020emergent}.

Some little accidents are indeed happy \citep{ross:book17}; PAIRED \citep{dennis2020emergent} proposes a novel, adversarial framework for simultaneously optimizing game-playing and game-building agents. Rather than implementing a filtering mechanism for candidate levels based on a minimal threshold of quality (i.e., the minimal criterion), PAIRED re-conceptualizes the notion as that of the Min player in a game of Minimax Regret \citep{dennis2020emergent, jiang2021prioritized, jiang2021replayguided, parkerholder2021that} but also incorrectly casts POET as a minimax algorithm. In \citet{openendedlearningteam2021openended} candidate environments are filtered based on satisfying a minimal threshold of quality similar to the minimal criterion described in POET. Like PAIRED, the feedback from multiple agents are considered when filtering candidate levels. However, code from \citep{openendedlearningteam2021openended} is not yet publicly available, and could prove susceptible to reinterpretations if it were.

This paper proposes a framework called Watts for designing reproducible and comparable experiments in open-ended learning. Motivated by modularity and algorithmic flexibility, Watts atomizes the components of OEL systems to promote the study of and the direct comparisons between OEL approaches. Grid-world learning environments can be developed and customized with the game-engine, Griddly \citep{bamford2020griddly}, designed with single and multiple agents in mind. Experiments can also be run on any of the learning environments already available in Open AI Gym \citep{brockman2016openai}. A distributed library for agent optimization called RLlib is used for single or multiagent learning \citep{rllib2018Liang}, which facilitates the design of custom algorithms or use of any of those currently available at \protect\url{https://docs.ray.io/en/latest/rllib-algorithms.html}\citep{rllib2018Liang}. While the innovation sought after in all of these open-ended learning paradigms can arise from cooperative or competitive interactions between agents and/or their environments, the challenge of open-ended learning lies in designing systems that can push search away from the boundaries of convergence toward areas of divergent intrinsic dynamics. To promote this research direction, we have open-sourced Watts and encourage the community to contribute.

To summarize our contributions:
\begin{itemize}
    \item We propose a modular framework - Watts - to build OEL algorithms with re-usable subcomponents.
    \item We implement POET and PAIRED in Watts and compare them with one another.
\end{itemize}

Section 2 details design considerations. Section 3 describes Watts' current modules. Section 4 describes how section 3's components can combine into OEL algorithms and how interpreting these pieces in different lights can result in related new algorithms. Section 5 compares these algorithms in a common suite of environments and section 6 discusses initial performance. Section 7 is results and future directions for Watts.

\section{Design Considerations and Motivations}
At the heart of popular frameworks like Open AI Gym is the notion that standardization enables good science \citep{brockman2016openai}. 
In software engineering and in particular in reinforcement learning,
encapsulation can promote code reuse and encourage readability such that errors in these complicated, stochastic, and distributed algorithms are quickly noticed and corrected \citep{rllib2018Liang}. 
One source of programmatic complexity in designing OEL systems is the large number of distinct modules with interlocking parts. When (co)evolutionary systems manage large populations of learners, they often have components for optimizing agents \citep{salimans2017:ESRL, schulman2017proximal}, generating environments \citep{Khalifa2020pcgrl}, evolving tasks dynamically \citep{openendedlearningteam2021openended}, filtering which candidate individuals can enter the population (i.e., minimal criterion \citep{brant:gecco17}), and transferring knowledge from a source learner to a target learner \citep{wang:gecco19}. Each of these components represent modules that are candidates for code reuse.

Good encapsulation can also promote good code structure, enabling a model of these algorithms as arrangements of their basic building blocks. Designing algorithms is an inherently creative process \citep{stanley:leo16}, and distilling these algorithms into their component blocks can enable combinatorial creativity \citep{Boden2009, NEURIPS2019PYTORCH}. Accessing multiple OEL approaches in a common framework helps illuminate the combinations of algorithmic elements yet to be imagined. Watts provides an abstraction that allows developers and researchers to focus on defining new modules without worrying about the complexity of the rest of the OEL algorithm. 



\section{The Watts Framework}
Watts is designed to conceptualize algorithms as modular pieces that can be mixed, matched, and recombined in novel ways. Shown in Figure \ref{fig:POET_ALG} are modularized components of OEL algorithms. Components for generating environments are called \textit{Generators}. Components that implement filtering are called \textit{Validators}. \textit{Solvers} define how agents are optimized in Watts. Components that interact with other elements of the outer loop are implemented as \textit{Strategies}, like an \textit{Evolutionary Strategy} or \textit{Transfer Strategy}. Such modularity enables a ``separation of responsibilities'' in implementation details to isolate hypotheses.

\subsection{Environment Generation}

Environments in Watts are created with \textit{Generators}. These generators are modules that enable any number of environment representations (e.g., indirect encodings, direct encodings, wave function collapse). Encapsulating environment generation to its own module enables a common API between new but algorithmically unrelated approaches to procedural content generation in OEL systems.

A generator that creates 2D grid-worlds could store direct (x, y) locations of the objects in the environment (e.g. player locations, wall tiles, traps, and treasure locations, etc.) and the rules that govern increase or decreasing amounts of these objects and/or rearrange their locations (e.g. remove a treasure; move the exit) -- this is how POET and PINSKY work \citep{wang:gecco19, dharna2020cogeneration}. When a generator is queried, a level is produced and variety is obtained by maintaining a population of distinct generators. 

Alternatively, generators could store neural networks that sequentially place objects on a blank canvas, such that a new level is created every time the generator is queried (e.g., a new maze ) -- this is how PAIRED and PCGRL handle level generation \citep{dennis2020emergent, Khalifa2020pcgrl}. And generators can operate on anything between these two extremes. A hypothetical generator from \citep{openendedlearningteam2021openended} could create a level topology with Wave Function Collapse \citep{Gumin_Wave_Function_Collapse_2016} and then query compositional pattern producing networks to place the remaining game-objects. In general, the environment generators can be thought of as parameterizing the missing information of an Underspecified Partially Observable Markov Decision Process (UPOMDP) \citep{dennis2020emergent}. 

\subsection{Evolution and Minimal Criteria}

An open-ended learning process should be able to shift the training distributions and objectives, and the field of evolutionary computation provides a natural approach to incrementally shifting distributions. 
The traditional view of an evolution strategy (ES) is one of being a zeroth order optimization algorithm for real-valued parameter-vectors. In an ES rather than calculate (stochastic) gradients of a loss function with respect to the model parameters, gradients are estimated from stochastic perturbations (sampled from an underlying noise distribution, usually a Normal) around the current parameter value. The key insight of this view is that the noise distribution itself can also be learned using feedback of the reward/fitness function e.g. CMA-ES \citep{Hansen06thecma}. Because an ES can contain a tunable representation of the underlying probability distribution, ES's are also generative algorithms.

An ES functioning as a generative algorithm provides a method of shifting agent training distributions (e.g. levels in a game). However, unlike using an ES to solve a black-box optimization problem, open-ended systems might require that the training distribution drift through parameter space without a target to optimize towards. Such schemes are called minimal criteria (MC) and represent an alternative to conventional fitness-based paradigms \citep{soros18Thesis} since the MC encodes a task-agnostic measure e.g. how similar is an agent to its neighbor \citep{Soros2016}. In Watts, MCs are represented as \emph{Validators}. Under this premise, so long as a minimum threshold on the quality of samples drawn from the training distribution is met, a candidate solution is accepted into the population of potential answers; the distribution does not change to explicitly optimize with respect to an evaluation metric.

MCs are of particular interest because they encode an undirected method of slow learning (i.e. evolution) to complement the fast learning of agent optimization \citep{botvinick2019fastandslow} resulting in easy to implement world-agent coevolutionary structures. To act as an outer-loop and shift populations, Watts provides an \emph{evolutionary} API. The evolutionary API accepts and returns a population, executing logic  that selects parents (i.e., select $k$ random parents; select the $k$ highest performers, etc.), mutates members of the population, and kills population members (i.e., kill the $k$ oldest members of the population if population size is exceeded). The evolutionary API can explore both traditional- and MC-based evolutionary structures as outer-loops on agent-learning. 

\subsection{Agent Learning}

In Watts the class that decouples the specific OEL implementations from the optimization algorithm is the \textit{Solver}. A Solver specifies the algorithm to optimize agent performance. The Solver has two functions that must be implemented: the optimize function and the evaluate function. The evaluate function returns the agent's fitness, and the optimize function defines a single step of optimization. For example, if the agents were optimized by David Ha's SimpleES~\citep{ha2017evolving}, the optimize function would first ask for potential solutions, evaluate them, return the answer to the optimizer and return the best agent found so far. However, there are other ways to define the optimization function as well. The default Single Agent Solver uses RLlib's \citep{rllib2018Liang} optimization algorithms (PPO, DDPG, OpenAI-ES, etc.).

Solvers are also our core unit of distribution for parallelizing work within Watts. Since OEL systems such as \citep{wang:gecco19, openendedlearningteam2021openended}, and \citep{dharna2020cogeneration} optimize a large population of agent/environment pairs, we define each Solver as a Ray Actor (\citep{Moritz2018RayAD}) so that each can optimize their models using separate cores, or even separate servers in a distributed setting. Providing a common distribution pattern allows researchers using Watts to ignore the complicated details of distributing work while keeping the performance benefits achieved through parallelization.

\subsection{Transfer Strategies}

One of the bottlenecks of algorithms like POET is that the transfer-learning step contains a combinatorial explosion of work to estimate a statistical ranking problem (i.e. what is the best agent for a given environment?). For increased code-modularity and re-use, Watts breaks down the ranking problem into two linked strategies: \textit{Scoring} and \textit{Ranking}. Scoring and ranking are separated to allow for careful study of how structuring the interaction (e.g., different types of ranking schemes) of individuals within a population affects learning \citep{jiang2009statistical, garnelo2021pick}. Separating ranking and scoring allows us to define what best means in multiple ways. For example, we could select the agent that scores the highest on a given task, or we could measure which environment induced the highest amount of information gain to pick a ``best'' agent even if it did not get the highest reward.

So far in POET-style algorithms, two ranking algorithms that have been considered. First is scoring the agents in zero-shot evaluations on every environment and selecting the single best agent for each environment \citep{dharna2020cogeneration}. A zero-shot evaluation scores an agent on a new environment without any additional optimization. The second ranking algorithm explored in POET is scoring zero-and-one-shot evaluations and selecting the single best agent for each environment \citep{wang:gecco19}.

For example, a zero-shot scoring strategy takes a Cartesian product of the current agents and environments returning a score for each agent in each environment (all evaluated in parallel). Then the rank strategy accepts the score matrix and takes an $argmax$ over the columns to pick new weights for each agent and environment pair. That is, the new weights for environment $k$ are picked via: $\argmax_{\theta}([R(\theta_1, E_k), ..., R(\theta_m, E_k)])$ where $R(\theta_m, E_k)$ is the zero-shot score of the $m$th agent on environment $k$.

Similarly, to implement the transfer strategy explored in \citep{wang:gecco19}, a new transfer strategy to pick weights for each agent-and-environment pair needs only wrap the zero-shot transfer strategy defined above and 1) perform a zero-shot evaluation 2) take one step of optimization for each agent 3) perform a one-shot evaluation and 4) take an argmax over the columns of the stacked evaluation matrices. That is, the new weights for environment $k$ are picked via: $\argmax_{\theta}([R(\theta_1, E_k), ..., R(\theta_m, E_k), R(\theta_1', E_k), ..., R(\theta_m', E_k)])$ where $R(\theta_m', E_k)$ is the score of the $m$th agent after one-step of optimization evaluated on the $k$th environment.

\section{Putting the pieces together} 

Figure \ref{fig:POET_ALG} shows an example of the \textit{Manager} class in Watts, which stores a population of agent-environment pairs. While the Manager class is itself algorithm agnostic, POET is diagrammed for the purpose of illustration. Implementation details for any given algorithm are specified in the component abstractions. 

\subsection{Diagramming POET}

\begin{figure}
    \centering
    \includegraphics[width=\textwidth]{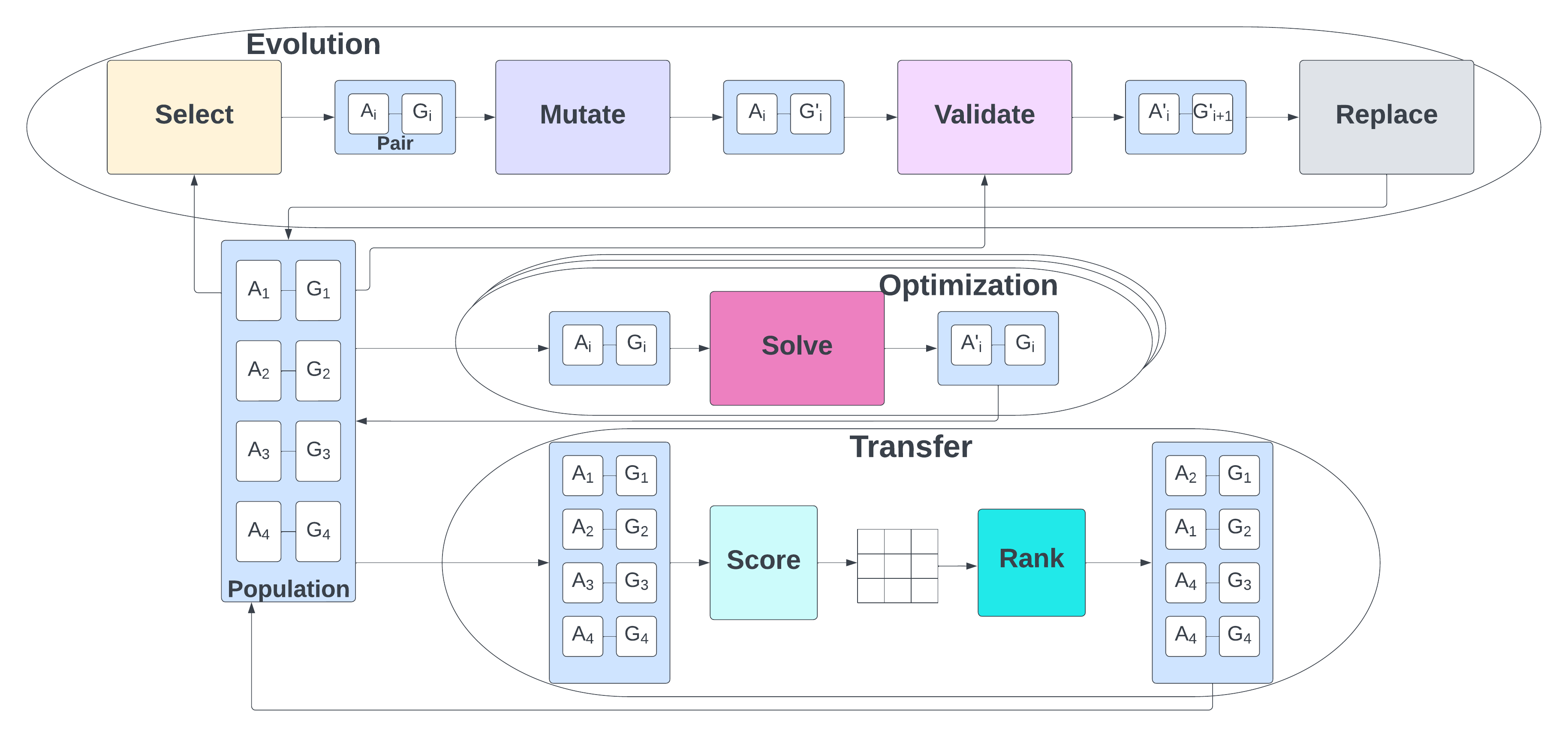}
    \caption{\textbf{Watts: The (POET)Manager Class} The POETManager stores an archive of agent and environment-generator pairs managed by Evolution, Optimization, and Transfer Components. Each of these components are further atomized and can be recombined to fit many different algorithms.}
    \label{fig:POET_ALG}
\end{figure}

The \textit{POETManager} is responsible for managing the inner-outer meta-learning loop of POET through components shown in Figure \ref{fig:POET_ALG}. The box of agent-environment pairs on the left is an archive of agent-environment pairs that represent the active population. The Evolution unit mutates the distribution of learning environments through a minimum criterion (MC)-based Evolution Strategy outlined as Evolution at the top of Figure \ref{fig:POET_ALG}. This collection of evolutionary components defines how to select parents from the meta-population, that are then evaluated by the Optimization components in the middle (i.e., a Solver). The Solver is where the algorithm for inner-loop learning is specified. The Transfer components at the bottom of Figure \ref{fig:POET_ALG} scores, ranks, and selects the best agents for a given environment (pseudocode shown in Appendix \ref{appendix:code}). So long as newly defined strategies (evolution, transfer, or optimization) follow their required APIs, new strategies can be slotted into the POETManager to experiment with creating new POET-like algorithms.

\subsection{PAIRED}

\begin{figure}
    \centering
    \includegraphics[width=0.75\textwidth, height=0.13\textheight]{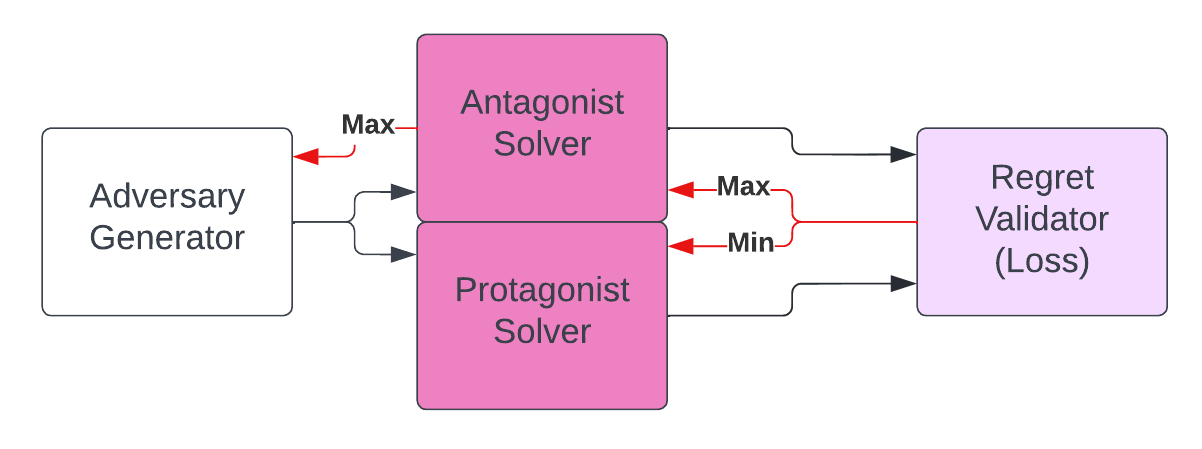}
    \caption{A potential diagram of PAIRED in Watts. In PAIRED, the adversary, antagonist, and protagonist are all parameterized by trainable neural networks and will require Validators and Generators to be differentiable objects. Black arrows are forward propagation and red arrows denote backwards gradient flow.}
    \label{fig:wattsPaired}
\end{figure}

While designed initially for POET-like algorithms, Watts also supports multi-agent reinforcement learning (MARL) \citep{dennis2020emergent} for open-ended learning.
Rather than keeping an archive of agent-environment pairs, PAIRED \citep{dennis2020emergent} trains a triple of agents to both build and solve 2-D mazes from the MiniGrid suite \citep{gym_minigrid}. Agent training is instead framed as an adversarial competition between a protagonist agent who wants to solve a maze and its asymmetric antagonist duo, which is an agent-environment pair. This adversarial duo includes a Generator artificial neural network (ANN), that takes feedback from its adversarial agent about the quality of the level that it has created. Together the goal is that the generator creates solvable levels as indicated by the performance of the antagonist agent, but that the protagonist will struggle to solve \citep{dennis2020emergent} shown in Figure \ref{fig:wattsPaired}. Formally, optimizing the minimax regret is performed by the adversary game-builder and the antagonist game-player who maximize the positive regret while the protagonist maximizes negative regret. For convenience, since the adversary and antagonist networks share a loss function, they will be denoted $\theta_{A}$ and the protagonist agent will be denoted $\theta_{P}$. The joint optimization of all three networks will continue until the networks arrive at a Nash Equilibrium -- a point in policy-space where no player has anything to gain by changing only their own strategy.

\subsection{Unifying POET \& PAIRED}

Like any stochastic based population training, the active population in POET is controlled by evolutionary mechanics. Specifically in POET parents are selected randomly from the archive and removed based on their age in the archive. The MC filtering evolves the population according to the following update rule: WLOG, select an environment, $E_i$ to mutate, $$E_{\text{child}} \sim E_{i}$$ $$MC := MC_{min} \leq R(\theta_{parent}, \text{E}_{child}) \leq MC_{max}$$ $$E_{i+1} \leftarrow \mathds{1}(MC).$$



Because the MC is an indicator function, it can only impart minimal geometric knowledge regarding how to move in the latent search space. However, PAIRED optimizes following the gradients of their formulation of the MC, where loss is calculated as the minimax regret between the antagonist and protagonist agents. Let $E_{\theta_{A}}$ be the environment created by the adversary. Then a regret-based MC is defined as: $$MC :=  R(\theta_{A}, E_{\theta_{A}}) - R(\theta_{P}, E_{\theta_{A}})$$ It follows then that agents following the loss function in PAIRED are following the directional gradients of the MC: $$\argmin_{\theta_{P}}\argmax_{\theta_{A}}(\nabla MC)$$ where $\theta_A$ seeks to maximize the regret signal while $\theta_P$ seeks to minimize the regret signal. Such formulation of the MC, while antithetical to the original formulation of an MC, may provide inspiration for new OEL designs and incorporating regret-based MCs into POET.

\section{Comparative Study}
\label{sec:compare}

To show how Watts can incorporate different algorithms, a POETManager runs the POET algorithm on the HardcoreBipedalWalker environment and the PINSKY algorithm on a port of the 2-D maze environment from the MiniGrid suite \citep{gym_minigrid} using the same Manager. Similar to the MiniGrid suite mazes, the maze environment shown is written in Griddly. The PAIRED algorithm is implemented in Watts and runs on the same port.

\begin{figure*}
    \centering
    \subfloat[Initial flat terrain that seeds  POET]{\includegraphics[width=\textwidth]{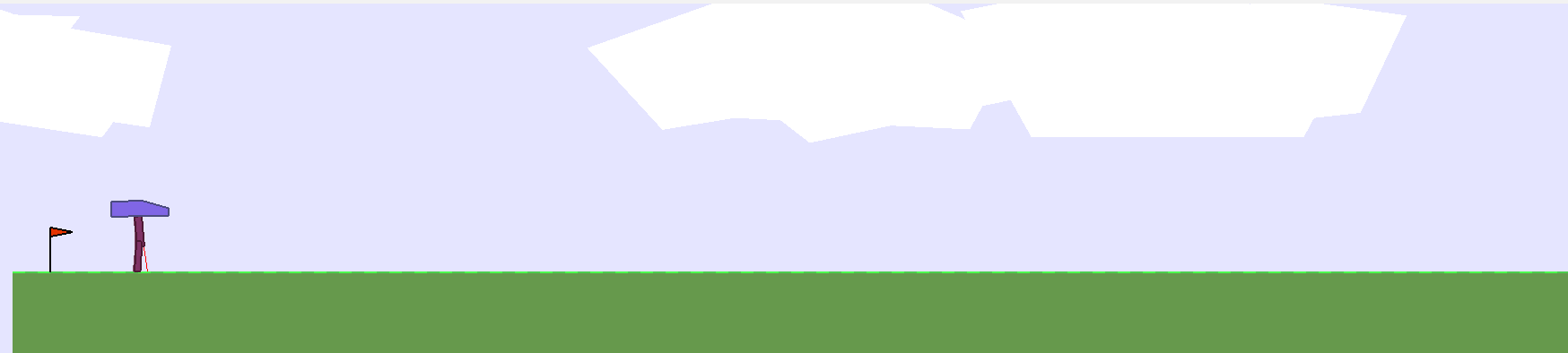} \label{fig:terrain0}}
    \hfill
    \subfloat[Example POET Generated Level]{\includegraphics[width=\textwidth]{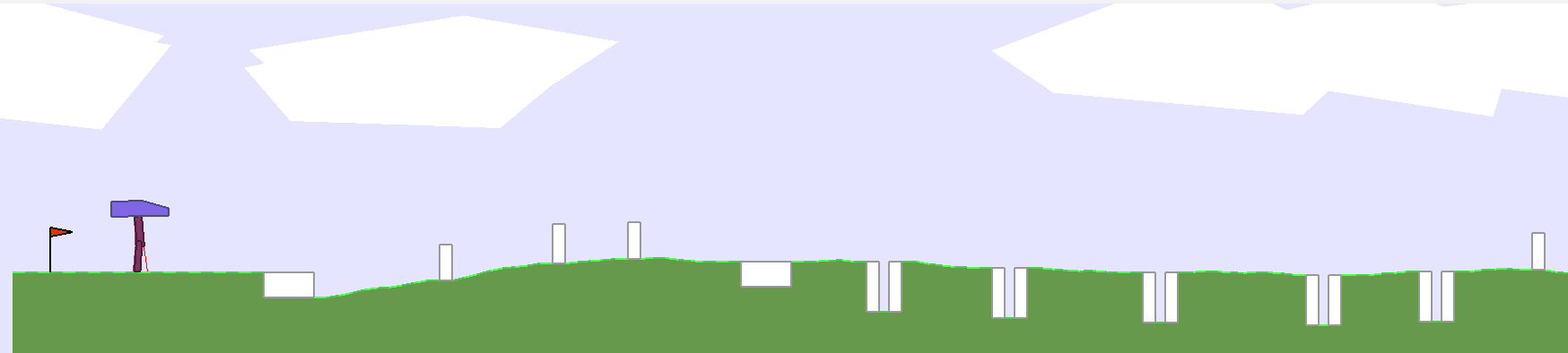}\label{fig:terrain143}}
    \caption{\textbf{Controllable Hardcore Bipedal Walker Environments integrated in Watts}. Bipedal agents in the Hardcore Bipedal Walker Environments start on the left hand side of the screen indicated by a flag shown in a) and b). While the terrain in a) is flat, the first obstacle the agent in b) encounters is a stair that it must step on or over to continue walking. The agent must then navigate over three stumps and a stair before encountering five pits and a stump at the rightmost portion of the environment.}
    \label{fig:terrains}
\end{figure*}

\subsection{Original POET in Walker and Maze}

The first experiment verifies that the POET-in-walker implementation is correct. As much of the walker-specific POET code is adapted from \citep{wang:gecco19}, it is unsurprising that the population behavior and generated terrains (Figure \ref{fig:terrains}) resemble those reported in \citep{wang:gecco19}. 
Interestingly when the terrain in the HardcoreBipedal walking environment was flat, a deterministic policy trained with the OpenAI-ES algorithm quickly converged to weights that resulted in rewards over the $MC_{max}$ threshold. However, the same policy did not score well when transferring to more complicated levels, struggling to perform better than  $MC_{min}$. Because the agent was rewarded without learning in its environment, the algorithm stopped challenging the agent with difficult levels it would normally add to the active population. However, training  stochastic policies for the walker restored the expected behavior. 

A useful experiment that can be run with Watts is to validate the assumptions made in \citep{dharna2020cogeneration}. When translating POET to work in the space of grid-worlds the authors made assumptions about how the sparsity of rewards in ``game''-style settings -- where agents are only rewarded for completing specific tasks (e.g. picking up a key, killing a monster) -- would be too sparse to allow for the MC first explored in POET to work, as the MC requires a well-aligned and reward-dense signal \citep{dharna2020cogeneration}. Therefore, by only changing the game environment and neural network structure, Watts can run the experiment of ``what would happen if $MC_{min} \leq R(\theta_{parent}, \text{E}_{child}) \leq MC_{max}$ from POET were used in the maze domain''? In the Maze domain, if the agent reaches the goal, the agent receives a reward of $R = 1 - (M/T)$, where M is the timestep on which the agent found the goal and T is the timestep horizon (500). In POET, the $MC_{min}$ and $MC_{max}$ are hyperparameters denoting an acceptable range of reward defining if levels are too-easy and too-hard. After 10,000 loops, Maze-based POET with $MC_{min}=0.1$ and $MC_{max}=0.9$ had a successful reproduction rate of $\frac{1}{800}$ with virtually all of the proposed levels being rejected due to being ``too hard'' because the agent scores a reward of less than 0.1 -- validating the assumptions. Such strict criteria (in the space of sparse-rewarding games) filtered out all chances for the coevolution to bootstrap itself as desired. 



\subsection{PAIRED in Maze}


Watts seeks to remedy the fact that until now attempts to compare POET against MARL algorithms have not been possible due to the high coupling of POET to the walker environment. To start this process, Watts provides environment wrappers that when combined with multi-agent learning algorithms in RLlib recreates the PAIRED algorithm defined in \citep{dennis2020emergent} for \emph{any} single-agent game that is defined in Griddly. 

As implemented in \citep{dennis2020emergent}, the adversary is constrained such that, at every time step, it chooses where to place a prespecified sequence of objects at available (x, y) locations. Relaxing the sequencing constraint by altering the Generator requires the adversary to also decide whether to place an object and if so to decide the type (i.e.,\ ``[place/not-place] [object type] at ([x], [y])''). With this loosened constraint and assuming that PAIRED converges to the Nash Equilibrium \citep{dennis2020emergent}, its learned behavior is characterized as a blank level without a goal, agent, or maze structure. Without these structural components, all agents receive zero reward. 


\section{Performance}
While performance is particularly important in OEL frameworks given the large amount of computational resources necessary to run experiments, often a significant portion of the distributed communication and execution code is rewritten when implementing algorithms for reinforcement learning \citep{rllib2018Liang}. 
For example, \citet{dharna2020cogeneration} implement custom code for distributed computing to run experiments on their POET-like algorithm, PINSKY \citep{dharna2020cogeneration}. Written in Python, experiments in grid-world environments completed a total of 5000 iterations in $\sim$2 weeks. On the other hand, the same experiments ran more quickly in Watts, with a total of 10000 iterations in $\sim$24 hours. Whether it is run locally or remotely on a cluster, computation in Watts is distributed through the Ray library \citep{Moritz2018RayAD}. Well-reasoned modules can promote comparison, and Watts implements some of the patterns common in distributed computing to ease implementing new OEL algorithms.

Interestingly, optimizing with the Open-AI ES \citep{salimans2017:ESRL} algorithm Watts runs about 10000 iterations in $\sim$3.5 days on 18 CPU cores. Optimizing the same HardcoreBipedalWalker environment, \citet{wang:gecco19} report 60,000 iterations in 12 days \citep{wang:gecco19, wang2020enhanced}) on 750 CPU cores. Shown in Table \ref{table:wattsPOETWallclock}, when Watts optimizes agent performance with PPO \citep{schulman2017proximal} it completes a total of 10000 POET iterations in $\sim$28 hours. It is hypothesized that Watts' ES POET runs more slowly than the original POET due to the number of compute cores the original algorithm used and additional experiments scaling up Watts are needed to test this. 


\begin{table}[ht]
\begin{center}
\begin{tabular}{||c c c c||} 
 \hline
  Algorithm & Implementation & Environment & $\frac{\text{loops}}{\text{cpu hr}}$ \\ [0.5ex] 
 \hline\hline
 Pinsky & Original & GVGAI & 0.47 \\ 
 Pinsky & Watts &  Griddly & 26.06 \\
 ES POET & Original & Walker & 0.28 \\
 ES POET & Watts & Walker & 6.61 \\
 PPO POET & Watts & Walker & 22.31 \\ [1ex] 
 \hline
\end{tabular}
\caption{Algorithm runtime comparisons.}
\label{table:wattsPOETWallclock}
\end{center}
\end{table}


\section{Discussion and Future Work}


The experiments show that we fulfilled our original design goal: an atomized framework for OEL algorithms such that only modular substitutions are needed to implement different algorithms. In Watts, POET and PINSKY only differ in how three modules are implemented. The Validator in POET requires that an agent performs within a given range of fitness on a candidate level. Whereas in PINSKY, the Validator checks for the existence of solutions to the level. For the Generator object, the environment encoding is changed based on the experimental domain. For PINSKY, the Ranking object performs zero-shot evaluations on the existing levels in the active population. Furthermore, the same POETManager can run on different domains with only changes to the Generator and representation of the agent in the Solver. In both cases only small implementation changes are required to run the algorithms in Watts. The goal of Watts is to explore the space of OEL algorithms, and Watts is a simple, flexible, and composable library. 

\begin{figure}
    \centering
    \includegraphics[width=\textwidth, height=0.4\textheight]{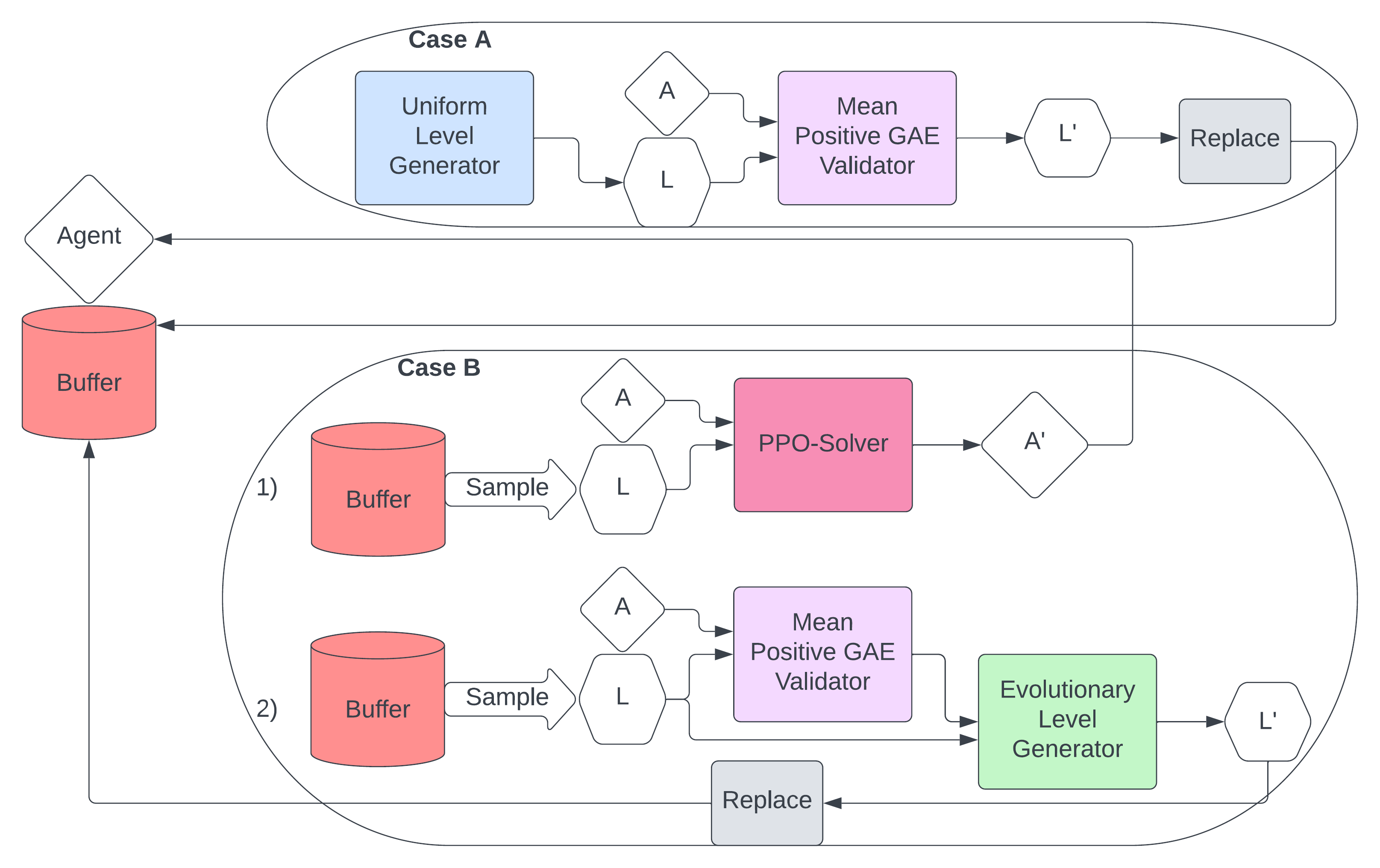}
    \caption{A hypothetical ACCELManager implemented out of Watts modules. For any given loop, sample case A or case B from a Bernoulli Random Variable. Case B is a two step process with one optimization step and one evolutionary-MC step. Note each individual component has been created just not strung together yet.}
    \label{fig:ACCEL}
\end{figure}

A consequence of modularity is illustrated in the head-to-head comparison of POET and PAIRED in Section \ref{sec:compare}. We provide a hypothetical Manager to implement the recent ACCEL algorithm \cite{parkerholder2021that} in Figure \ref{fig:ACCEL}. To improve the framework and its contribution to the community, we plan to add components for more OEL algorithms \citep{jiang2021prioritized,parkerholder2021that,bontrager2021learning,openendedlearningteam2021openended}. Interestingly, even just thinking about running PAIRED in the HardcoreBipedalWalker environment inspires the notion of differentiable Generators and Validators. Furthermore, something like a Generator API could provide more functionality, like the ability to sequentially build levels. With more functionality, the aim is to build an active core community of people interested in all aspects of OEL (including procedurally generated content).

Going beyond the comparison of existing algorithms, Watts can potentially enable the invention of new and more capable open-ended learning methods. One aim of Watts is to enable the easy testing of new algorithmic ideas through its modularity, like the recent population-based extension to PAIRED \citep{du2022takes} that pushes the algorithm even closer to POET. Pushing POET more toward PAIRED, we aim to explore incorporating a multi-agent minimax regret signal like that in PAIRED. Both innovations to these algorithms are conceivable in the Watts framework.  

Finally, the goal of Watts is to explore the space of OEL algorithms therefore designing simple, flexible, and composable algorithmic interfaces are a must; in order for others to use Watts, we must have documentation instructing how to setup and easily use the system. It must be able to be run on a variety of operating systems without performance disparity. And there must be a community of active users and contributors guiding and encouraging any newcomers.

\section{Conclusions}
We presented Watts, a framework for comparing existing open-ended learning methods and inventing new ones. In the experiments described here, we implemented the well-known POET and PAIRED algorithms and compared them, showing that they are each dependent on the particular problem domain they are applied to. We believe that breaking existing methods into their constituent components and recomposing them in a common framework will enable us and others to more rapidly iterate on novel OEL methods.

\bibliography{iclr2022_conference}
\bibliographystyle{iclr2022_conference}

\appendix
\section{Appendix}
\subsection{Pseudocode}
\label{appendix:code}

\begin{figure}[h]
    \centering
    \includegraphics[width=\textwidth, height=0.5\textheight]{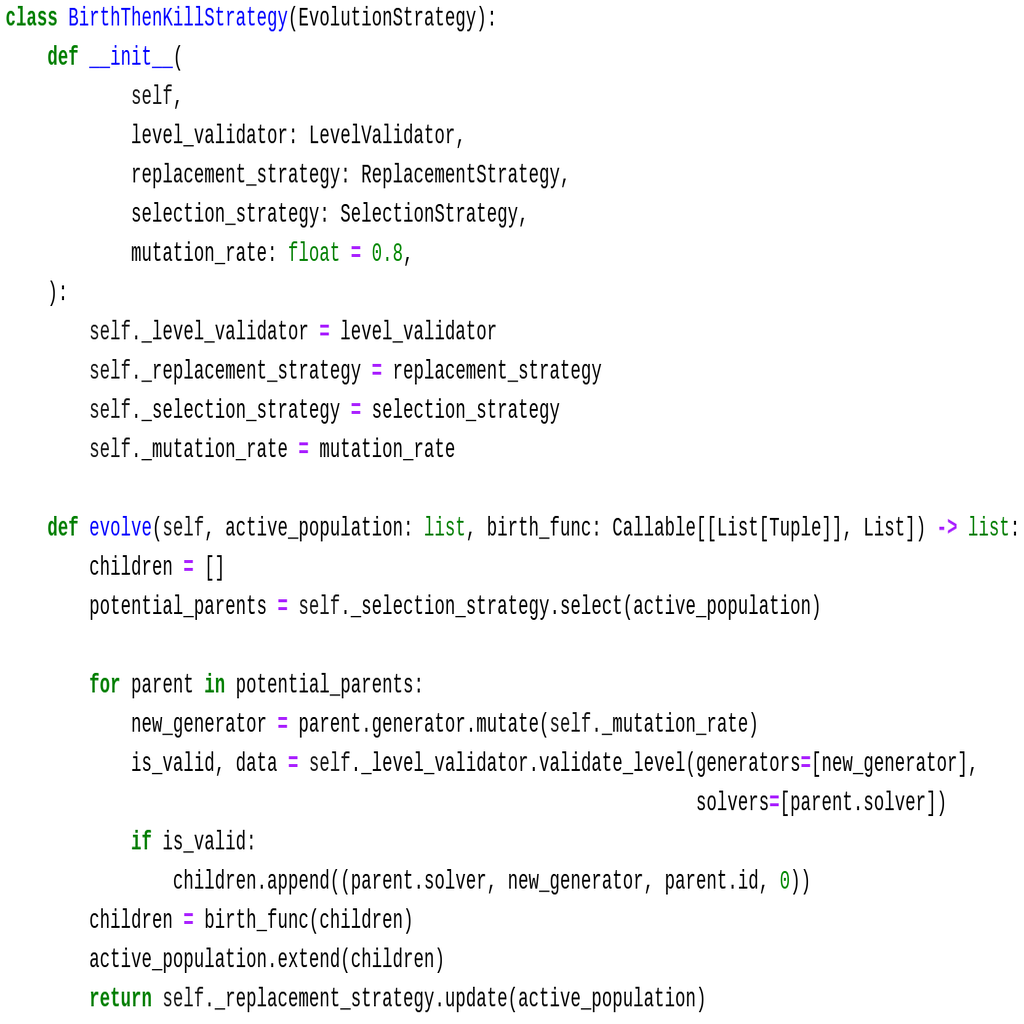}
    \caption{Evolution Strategy Pseudocode for an MC-guided Outer-Loop}
    \label{fig:ES}
\end{figure}

\begin{figure}
    \centering
    \includegraphics[width=0.7\textwidth, height=0.3\textheight]{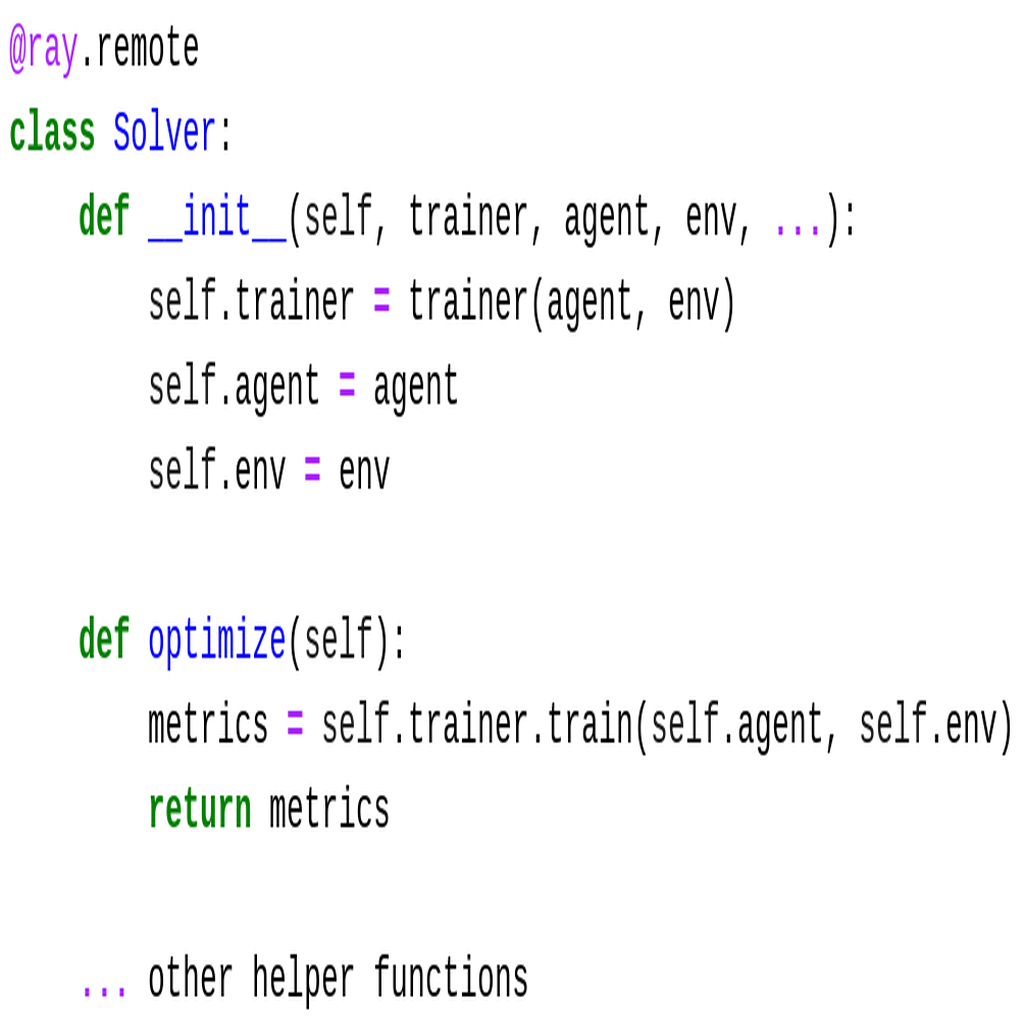}
    \caption{Example Solver that implements an optimization algorithm (trainer) on a remote process.}
    \label{fig:solver}
\end{figure}

\begin{figure}
    \centering
    \includegraphics[width=\textwidth, height=0.75\textheight]{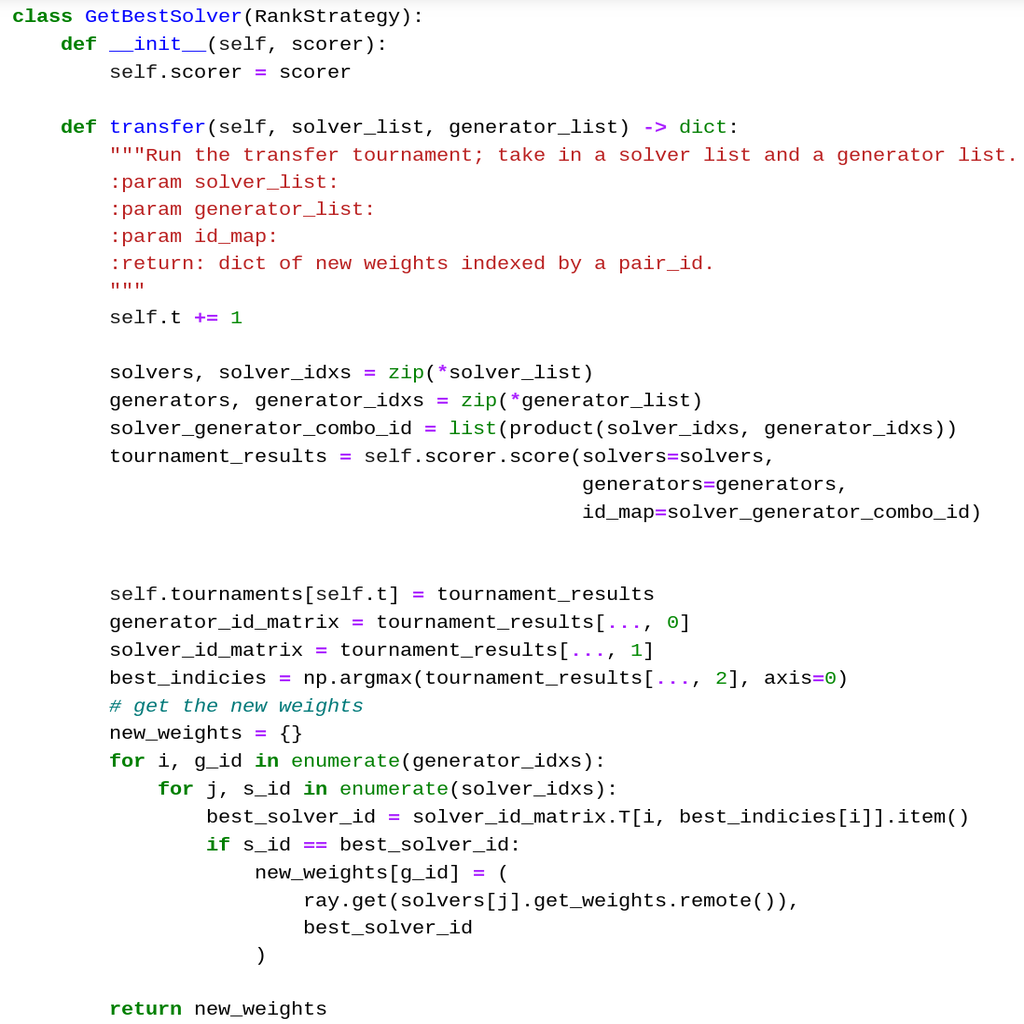}
    \caption{Transfer Strategy Pseudocode: Picking the best zero-shot agent for each env}
    \label{fig:transfer}
\end{figure}

\begin{figure}
    \centering
    \includegraphics[width=\textwidth, height=0.75\textheight]{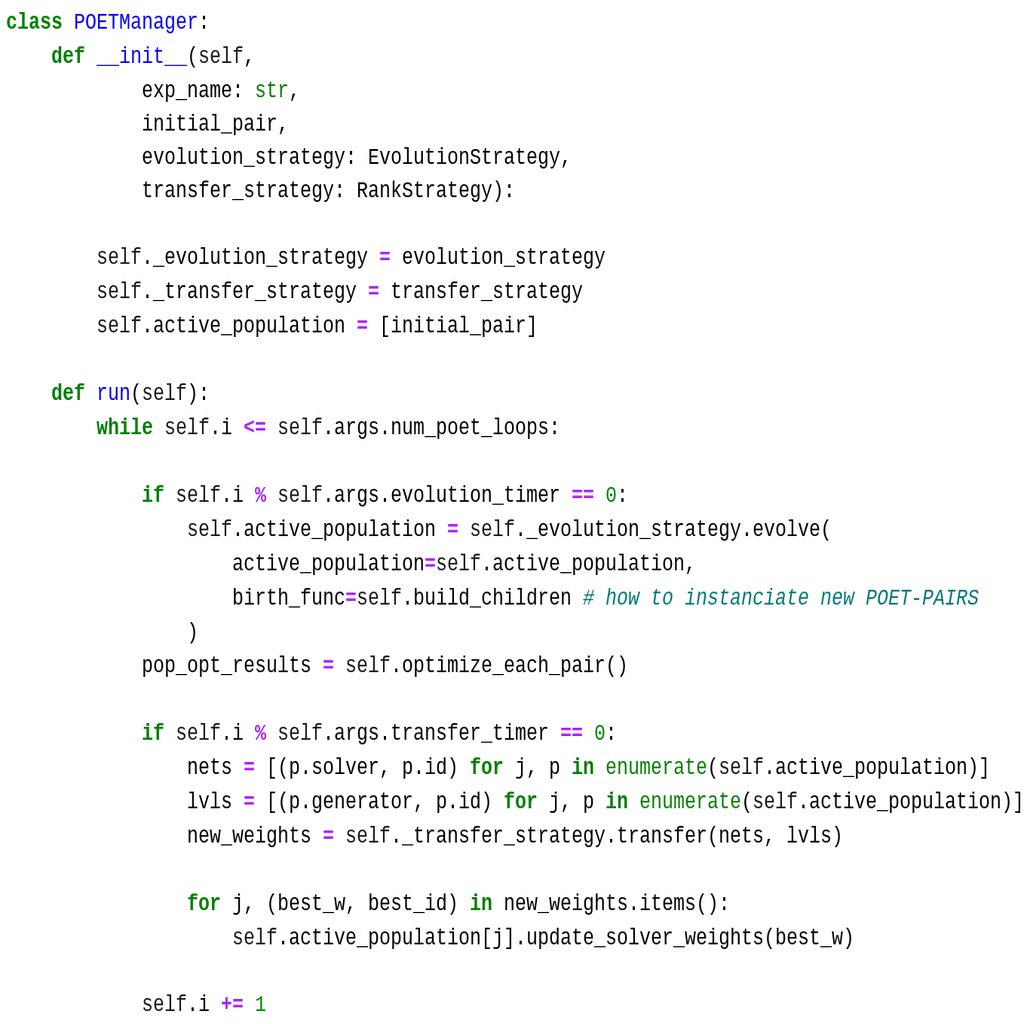}
    \caption{The POET algorithm modular form}
    \label{fig:poet}
\end{figure}

\end{document}